\documentclass[11pt]{article}

\usepackage[utf8]{inputenc}
\usepackage[T1]{fontenc}
\usepackage{lmodern}
\usepackage[margin=1in]{geometry}
\usepackage{graphicx}
\usepackage{bm}
\usepackage{float}
\usepackage{amsmath,amssymb}
\usepackage{booktabs}        
\usepackage{authblk}         
\usepackage[hidelinks]{hyperref}
\usepackage{cite}            

\title{Learning from Lost Provenance: Multiple Instance Learning\\
       for Cancer Registry Tumor Group Classification}

\author[1]{Leonard Ruocco}
\author[1,2]{Jonathan Simkin}
\author[2]{Lovedeep Gondara}
\author[3]{Gregory Arbour}
\author[3]{Raymond Ng}
\affil[1]{British Columbia Cancer Registry, Provincial Health Services Authority, Vancouver, Canada}
\affil[2]{University of British Columbia, Vancouver, Canada}
\affil[3]{Data Science Institute, University of British Columbia, Vancouver, Canada}

\date{\today}

\begin{document}

\maketitle

\abstract{
Modernizing cancer registries with deep learning is opening new opportunities to automate labor-intensive tasks such as the coding of pathology reports. However, progress is constrained by the scarcity of report-level human-annotated training data. Cancer registries generate substantial volumes of expert-assigned labels as a routine product of their operations, but these exist at the patient level and are not linked to the individual pathology reports that informed them, limiting their direct use for training models. We develop an efficient framework for training deep learning classifiers by leveraging these operationally-generated labels without requiring per-report human annotation, demonstrated for tumor group classification at the BC Cancer Registry. We use Attention-Based Multiple Instance Learning (ABMIL) to recover the lost link between patient-level labels and the reports that informed them, leveraging the attention the model places on each report to distil a large, noisily-labeled corpus into a compact, high-quality per-report training dataset. A classifier fine-tuned on a distilled dataset achieved a macro F1 of 0.83, outperforming established baselines across most tumor groups. By turning routine operational labels into high-quality training data without additional annotation or large-scale computing infrastructure, ABMIL offers a practical and accessible route to automating cancer registry workflows.
}

\section{Introduction}

Population-based cancer registries (PBCRs) play a foundational role in cancer surveillance, informing public health policy, resource allocation, and epidemiological research \cite{parkin2006evolution, blumenthal2020using}. A core operational task in any PBCR is the assignment of tumor group — a coded classification of cancer type used to direct cases to the appropriate tumor-specific coding workflows and populate surveillance databases\cite{parkin2006evolution}. At the BC Cancer Registry (BCCR), one of Canada's largest provincial cancer registries, this task involves classifying incoming pathology reports into one of many standardized tumor groups spanning the full spectrum of malignancy. The volume of incoming reports is substantial: BCCR receives approximately 100,000 reportable pathology reports annually, and the manual effort required to assign tumor group classifications represents a significant and growing burden \cite{mousavi2025challenges, merriman2021evolution} on certified Oncology Data Specialists (ODS)---trained health information management professionals who code cancer cases into registry databases. 

Natural language processing (NLP) offers a principled route to automating classification of pathology reports \cite{lopez2022natural}. Pathology reports are semi-structured free-text documents with a consistent clinical vocabulary and relatively constrained linguistic variation, making them well-suited to supervised text classification \cite{hands2025survey}. Recently, deep learning approaches have supplanted traditional rule-based and statistical machine learning (ML) methods for clinical NLP tasks \cite{min2023recent}, with transformer-based models now representing the state of the art \cite{vaswani2017attention}. Central to these approaches is the representation of text as dense embedding vectors that capture the semantic content and contextual relationships between words \cite{devlin2019bert}. 

A central challenge in developing NLP systems for cancer registry automation is the scarcity of pathology reports labeled at the individual document level in a format suitable for supervised ML. Constructing such datasets requires substantial subject matter expert involvement; as an ODS must review and label each report. This resource has been consistently identified across the clinical NLP literature as a primary bottleneck to the development and scaling of supervised models \cite{de2021deep, santos2022automatic}.

A natural alternative is to leverage the large body of labels that ODSs already generate as a routine product of their daily workflow. During normal registry operations, ODSs review incoming pathology reports, abstract relevant clinical information, and record tumor group assignments in the registry database. The obstacle is structural: registry labels are recorded at the tumor level and are the consolidated output of reviewing one or more pathology reports alongside other clinical documentation. However, the link between a given label and the specific reports that informed it is not recorded.

The standard approach in the literature has been to filter the data for patients with single reports \cite{de2021deep}, or to propagate the patient-level label to every pathology report associated with that patient and treat the resulting pairs (report, label) as independently labeled training examples \cite{chandrashekar2024path, gao2019classifying}. Filtering to single-report cases avoids label noise but discards much of the available data and biases the sample toward non-representative patients. Label propagation avoids this data loss but is misspecified: it assigns the tumor group label to all reports for a patient regardless of clinical relevance, including workup biopsies, incidental findings, and reports from anatomical sites unrelated to the registered tumor. The resulting training signal degradation is well documented \cite{gao2020using} and is particularly damaging for minority tumor groups where clean examples are already scarce.

Another proposed solution is to concatenate all reports into a single block of text and rely on the model to identify the relevant signal within it \cite{preston2023toward}. This introduces a practical limitation due to the length of concatenated text requiring long context windows, which can quickly exceed computational constraints for large datasets. Furthermore, dissolving reports into a single sequence removes the document as a discrete unit. Given that many reports in a patient bag are clinically irrelevant to the registered diagnosis, any method that operates over the full concatenated sequence cannot selectively exclude them. Retaining the report as an isolable unit opens the door to dataset distillation: identifying and discarding low-relevance reports prior to or during training, which can dramatically reduce corpus size and computational burden without sacrificing the diagnostic signal. 

We propose that all these challenges are jointly addressed by a single well-established framework: Attention-Based Multiple Instance Learning (ABMIL). In the ABMIL framework \cite{ilse2018attention}, training labels are assigned at the level of a `bag' of `instances' rather than at the individual instance level. A bag is labeled positive if at least one of its instances is positive; the model learns to identify which instances are responsible for the bag-level label without requiring instance-level annotations. This framework has been applied analogously in computational pathology, where histopathology images are treated as bags of patches supervised by slide or report derived labels \cite{campanella2019clinical, marini2022unleashing}; here we apply the same principle to text, treating patients as bags of pathology reports supervised by registry labels. Unlike label propagation, the model is never penalized for any individual report failing to predict the bag label. Supervision is applied only at the level of the aggregated bag representation, leaving the model free to down-weight irrelevant reports without gradient penalty. This matches the BCCR setting precisely: each patient health number (PHN) with a single registered tumor constitutes a bag, the pathology reports associated with that PHN are the instances, and the BCCR record tumor group is the bag label. The model is never asked to predict the label from any individual report; it is asked only to aggregate evidence across a patient's full set of reports and predict the tumor group at the tumor level.

Beyond its use as a standalone model, the ABMIL framework offers a second operational pathway of practical value to PBCRs. These per-report relevance scores can be used to filter the full registry-linked corpus, retaining only the highest-scoring reports per PHN as a proxy for the diagnostically relevant documents. The tumor group label is then propagated to all surviving reports, yielding a filtered dataset in which every retained report carries the PHN-level tumor group label. This is a clean one-to-one mapping structurally equivalent to a conventionally annotated per-report dataset, suitable for training a standard supervised classifier. This two-stage pipeline therefore offers a route to constructing large, high-quality training corpora from operational registry data without any subject matter expert annotation effort, with the relevance scores serving as an automated document selection step that would otherwise require manual review. In this paper, we evaluate both pathways: direct classification using the trained ABMIL model, and relevance-guided dataset filtering followed by supervised fine-tuning of a downstream classifier. To our knowledge, no prior published system has applied ABMIL to cancer-registry tumor-group classification over free-text pathology reports with operationally-generated bag labels.

Beyond the BCCR, the proposed approach has broad applicability: virtually all PBCRs face the same structural challenge of patient-level labels with no recorded link to individual reports, and any registry with an operational coding database and linked report repository could in principle apply this method. Because this filtering step distils the corpus to its most diagnostically relevant reports, model training remains computationally modest — positioning the approach as a practical framework for democratising cancer registry NLP across institutions of varying resource capacity.

\section{Data Sources}

Two operational databases maintained by the BCCR were used in this study. The first is the BCCR pathology report database, which stores free-text surgical pathology reports submitted to BCCR from laboratories across British Columbia. Each report is identified by a unique message identifier and linked to a patient via their PHN. The second is the BCCR tumor database which includes clinical information on all reportable tumors among British Columbians fully abstracted by ODSs according to coding standards \cite{icdo3}. Tumor group assignments in this database are derived using the CAIS mapping — a site-code and histology-code to tumor group mapping schema used at BCCR — which serves as the source of bag-level supervision labels throughout this study. The two databases were linked via PHN to produce a \textit{corpus} in which each patient is associated with both a set of pathology reports and BCCR tumor database derived tumor group label (see Table \ref{tumor_distribution} for the set of tumor groups used in this study and the corresponding numbers of reports extracted for each group). Data were extracted for the period 2020-01-01 to 2023-12-31, yielding 82,855 unique PHNs comprising 359,714 pathology reports.

\begin{table}[h]
\centering
\begin{tabular}{l l r r r r r}
\hline
\textbf{Code} & \textbf{Tumor Group} & \textbf{N PHNs (\%)} & \textbf{N Reports (\%)} & \textbf{Median Bag Size} & \textbf{Mean Bag Size} \\
\hline
GU & Genitourinary    & 16657 (20 & 60497 (17) & 3 & 3.6 \\
GI & Gastrointestinal & 15900 (19) & 58734 (16) & 3 & 3.7 \\
BR & Breast           & 11656 (14 & 55916 (15) & 4 & 4.8 \\
LU & Lung             & 8403  (10) & 37177 (10) & 4 & 4.4 \\
ME & Melanoma         & 6130  (7)  & 30586 (9)  & 4 & 5.0 \\
GY & Gynaecological   & 5766  (7)  & 26497 (7)  & 4 & 4.6 \\
LY & Lymphoma         & 5038  (6)  & 33158 (9)  & 6 & 6.6 \\
HN & Head and Neck    & 4101  (5)  & 16682 (5)  & 3 & 4.1 \\
LK & Leukemia         & 3383  (4)  & 19741 (6)  & 4 & 5.8 \\
SA & Sarcoma          & 2570  (3)  & 9424  (3  & 3 & 3.7 \\
NE & Neuroendocrine   & 2045  (3)  & 5048  (1)  & 2 & 2.5 \\
SK & Skin             & 1206  (2)  & 6254  (2)  & 4 & 5.2 \\
\hline
\textbf{TOTAL} &  & \textbf{82855} & \textbf{359714} & \textbf{4} & \textbf{4.3} \\
\hline
\end{tabular}
\caption{Distribution of tumor groups and bag statistics across the dataset.}
\label{tumor_distribution}
\end{table}

PHNs with more than one unique tumor diagnosis recorded in the BCCR tumor database were excluded. This restriction is necessary to ensure an unambiguous bag-level label: a patient with two distinct primary tumors may have pathology reports associated with either diagnosis. The dataset was partitioned into training and validation sets using patient-stratified splitting, ensuring that all reports belonging to a given patient appear exclusively in one partition preventing patient-level data leakage across splits \cite{rosenblatt2024data}. The final split comprised 90\% training and  10\% validation PHNs. 

The validation set was used to track training performance metrics and allow each model to reach convergence. Final evaluation and methodology comparison were performed on an independent dataset of 2,057 pathology reports collected during 2025. We refer to this as the \textit{ODS-annotated dataset}. Because report-level gold labels are independent of the registry propagation process, this dataset serves as our primary benchmark for assessing model performance under operationally realistic conditions. Labels were assigned using a SEER-based site-code and histology-code to tumor group mapping, which differs from the CAIS mapping used to derive training labels from the BCCR tumor database. While the two mappings largely align, a small proportion of cases receive discordant tumor group assignments under the two schemas. This introduces a modest degree of label noise at test time that is independent of model quality and should be accounted for when interpreting absolute performance figures on this dataset.

\section{Methods}

This pipeline leverages the ABMIL framework\cite{ilse2018attention}, formally defined as follows. Given a bag of $N$ report embeddings $\{h_1, ..., h_N\}$, where each $h_i \in \mathbb{R}^d$ is a dense vector representation of a single pathology report extracted from the final hidden layer of a pretrained encoder with hidden dimension $d$, the model computes a gated attention weight $a_i$ for each instance as:
\begin{equation}
a_i = \frac{\exp\left(w^T \left(\tanh(V h_i) \odot \sigma(U h_i)\right)\right)}{\sum_j \exp\left(w^T \left(\tanh(V h_j) \odot \sigma(U h_j)\right)\right)}
\label{attention_weights}
\end{equation}
Here $V \in \mathbb{R}^{L \times d}, \quad U \in \mathbb{R}^{L \times d}, \quad w \in \mathbb{R}^{L \times 1}$ are learnable parameters and $L$ is the attention hidden dimension. The resulting attention weights are used to compute a bag-level representation $z=\sum_{i=1}^{N} a_i h_i \in \mathbb{R}^d$, which pools the instance embeddings into a single fixed-size vector weighted by each report's estimated relevance to the bag label. This bag embedding is then passed through a fully connected classification head with softmax activation over the 12 tumor group classes.

\begin{figure*}[t]
\centering
\includegraphics[width=0.8\textwidth]{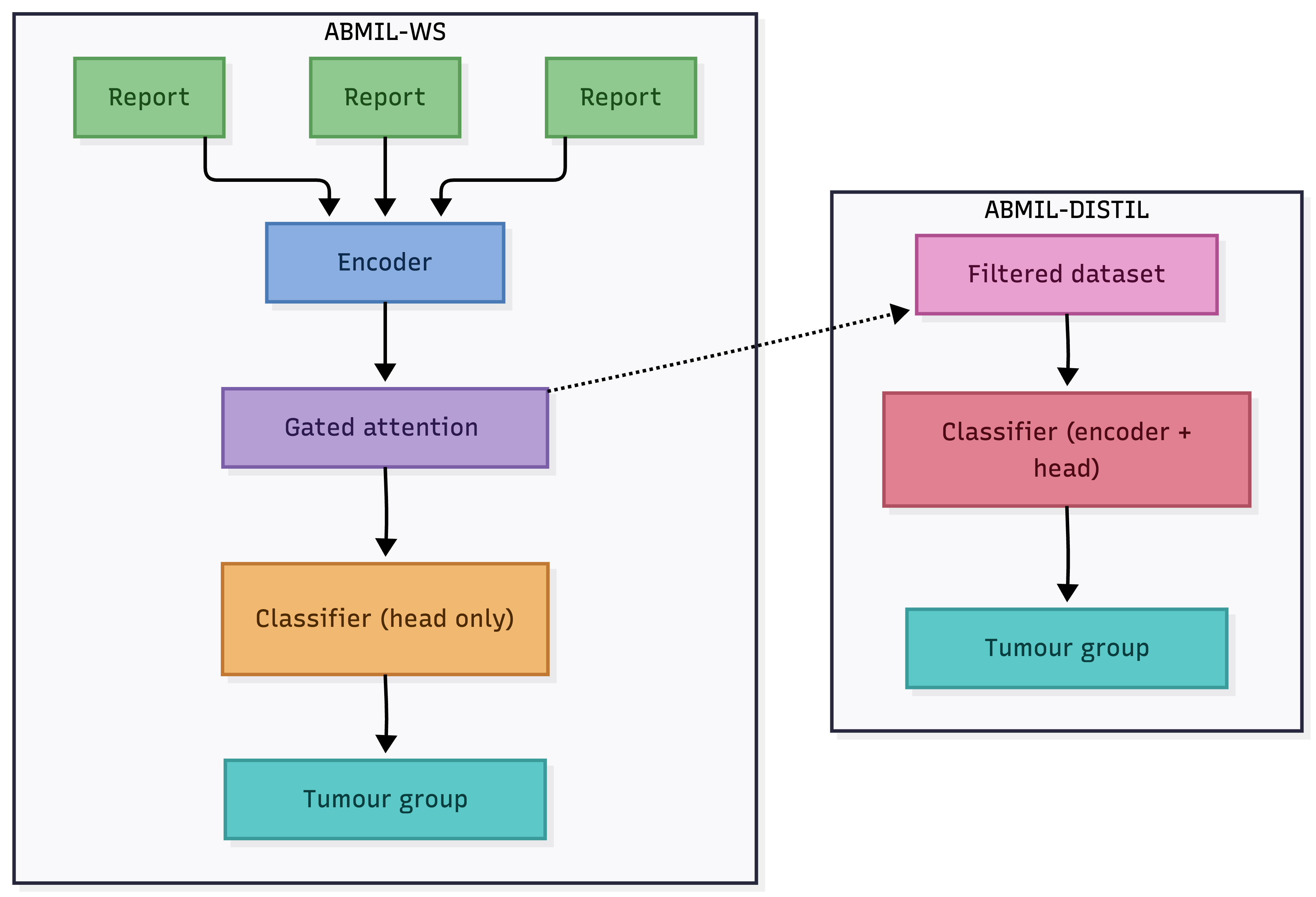}
\caption{Schematic of ABMIL pipeline. The first pathway ABMIL-WS refers to the weak-supervision classification method, where ABMIL is used to train a classification head on frozen embeddings of bags of pathology reports. The second pathway, dubbed ABMIL-DISTIL uses the attention scores established from ABMIL-WS and filters the corpus to produce a distilled dataset which is passed to a classifier for full end-to-end fine-tuning.}
		\label{ABMIL_pipeline}
\end{figure*}

The proposed method consists of a two stage pipeline indicated in Figure \ref{ABMIL_pipeline}. In the first, the ABMIL framework is used directly as a weakly supervised classifier trained on bags of reports. We refer to this section as ABMIL-WS. In the second, the ABMIL filtering pathway, the attention weights produced by the trained ABMIL-WS model are repurposed as per-report relevance scores to filter the \textit{corpus}, after which a classifier is trained on the resulting high-confidence dataset, with gradients free to backpropagate into and update the encoder weights. This stage is referred to as ABMIL-DISTIL. 

Each pathology report was independently encoded to a fixed embedding using BCCRTron \cite{gondara2024classifying}, a domain-adapted clinical language model produced by continued pretraining \cite{gururangan2020don} of GatorTron-base \cite{yang2022large} on approximately 1.1 million BCCR oncology pathology reports. GatorTron-base is a 350M parameter clinical language model pretrained on over 90 billion words of clinical text, and BCCRTron inherits its architecture and parameter count. For each report, the [CLS] token — a fixed summary representation produced by the encoder for each input sequence \cite{devlin2019bert} — was used as the report-level embedding, yielding a vector of dimension d=1024 with a context window of 512 tokens.

In the ABMIL-WS stage of the pipeline, the pretrained encoder (BCCRTron) is used as a fixed feature extractor: each report in a PHN bag is encoded once to its [CLS] embedding, and only the gated attention pooling layer and the classification head (see Figure \ref{ABMIL_pipeline}) are trained, using the bag-level BCCR tumor group labels as the supervision signal. The gated attention network projects each report embedding through a hidden layer of dimension 512 to compute its attention weight (Equation \ref{attention_weights}); the weighted embeddings are then pooled into a single bag-level representation and passed to the classification head to predict the tumor group. Because no gradients propagate into the encoder, training updates only the small attention and classification parameters — a lightweight procedure that requires no encoder fine-tuning and operates entirely on the precomputed embeddings.

For the ABMIL-DISTIL stage, the attention weights learned by the ABMIL-WS model are used to filter the \textit{corpus} into a series of progressively higher-quality (but lower $N$) training datasets. For each PHN in the training set, reports are ranked by their attention score $a_i$ and a percentile-based threshold is applied to the distribution of $a_i$ across all reports in the corpus. Reports falling above the threshold are retained as high-confidence representative documents for their respective PHN; reports below the threshold are discarded. 

This filtering is applied at multiple percentile thresholds (i.e. 0, 50, 75, 85, 90, 95, 99 percentiles), yielding datasets of varying size. As the threshold increases, the retained dataset size decreases but each report is more likely to be the diagnostically relevant document that informed the ODS coding decision. However, the trade-off is that for higher thresholds, less reports pass the filter and the resulting training dataset size is lower which can effect model performance; we explore this trade-off in our results. At a threshold of 0\%, all reports are retained regardless of attention score, reproducing the label-propagation setup exactly (referenced in the Introduction).

A supervised classifier is then independently trained on each filtered dataset using BCCRTron as the encoder backbone. The primary motivation for the filtering approach is that reducing the corpus to high-attention reports makes end-to-end fine-tuning computationally tractable, a key practical advantage over training on the full unfiltered corpus. Across the threshold sweep, we use a partially frozen encoder with 6 of 24 transformer layers unfrozen, as training fully end-to-end across all threshold conditions was not feasible within available compute constraints. For model evaluation and threshold selection, the \textit{ODS-annotated dataset} was split in two: the filtering threshold was selected on one half (dev) and performance reported on the other (test), ensuring no report contributed to both threshold selection and the reported performance estimate.

\subsection{Baseline Comparisons}
Two baselines are evaluated alongside the proposed approach. The first is the \textit{label-propagation baseline}. This baseline is equivalent to the 0\% attention threshold case in the filtering experiment: the ODS assigned tumor group label for each PHN is propagated to every associated pathology report, and a supervised classifier is trained on the resulting per-report dataset. Every report in the corpus is retained regardless of its relevance to the registered diagnosis, and each is treated as an independently labeled training example. 

The second baseline is the concatenated-bag baseline. Here all pathology reports associated with a given PHN are concatenated into a single block of text and the PHN-level tumor group label is assigned to the resulting sequence. To accommodate the greater length of concatenated bags, we use BCCRTron-2k as the base model for this classifier, which supports a larger context window (2048 tokens) than the BCCRTron variant used in the main experiments (512 tokens). BCCRTron-2k is itself a continued-pretraining of GatorTron-base-2k and shares the same parameter count and weights, differing only in context window size. For both baselines, the same encoder layers were trained and the same hyperparameters used (see Supplementary Material).

\section{Results}

We now present results of running the full pipeline outlined in Methods. The ABMIL-WS stage is run first: encoding all reports and training a small attention pooling and classification head using bag-level supervision. Attention scores are bounded between 0 and 1, with the majority of reports receiving weights close to 0 and a long right tail — reflecting the model's tendency to concentrate predictive signal on a small subset of reports per bag (see distribution in Supplementary Material). This is consistent with the assumption that a minority of reports carry the diagnostic signal, and justifies using attention percentiles as a filtering criterion. Retention rates across tumor groups and their relationship to bag size are reported in the Supplementary Material.

\begin{figure*}[t]
\centering
\includegraphics[width=0.9\textwidth]{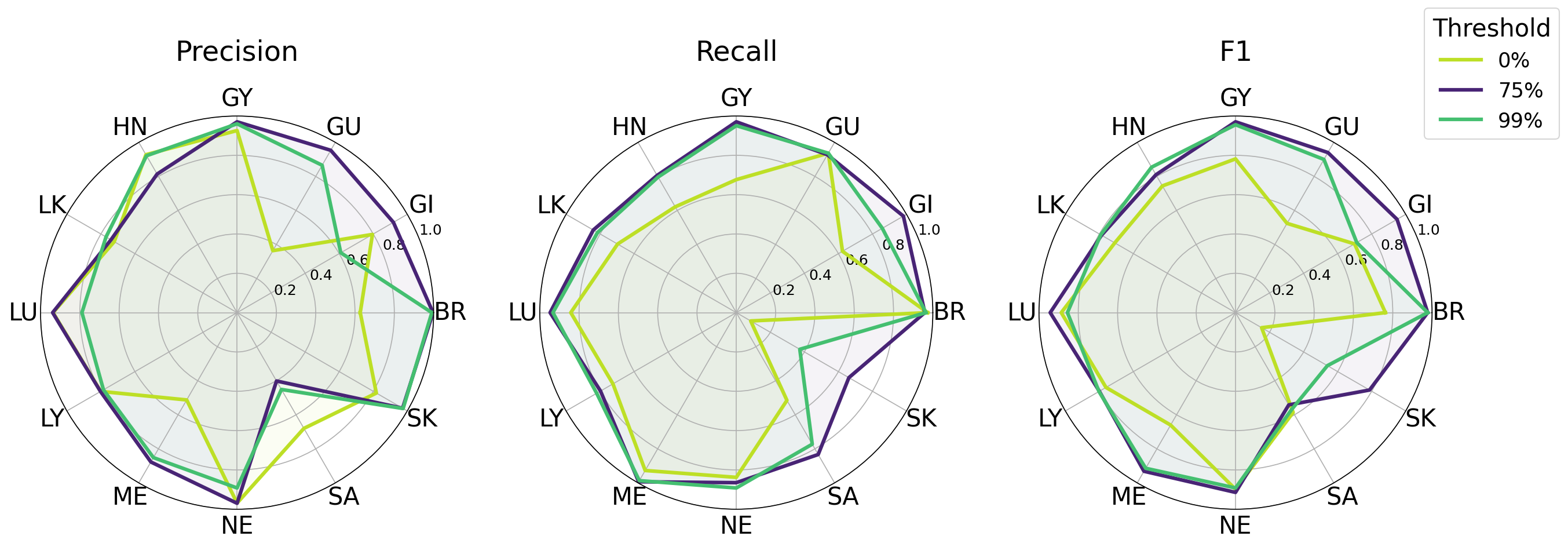}
\caption{Per-class precision, recall, and F1 for the ABMIL-DISTIL supervised classifier at three attention score percentile thresholds: 0\% (label-propagation baseline; yellow), 75\% (optimal; purple), and 99\% (maximum filtering; green).}
		\label{radar_threshold_comparison}
\end{figure*}

\begin{figure}[t]
\centering
\includegraphics[width=0.5\textwidth]{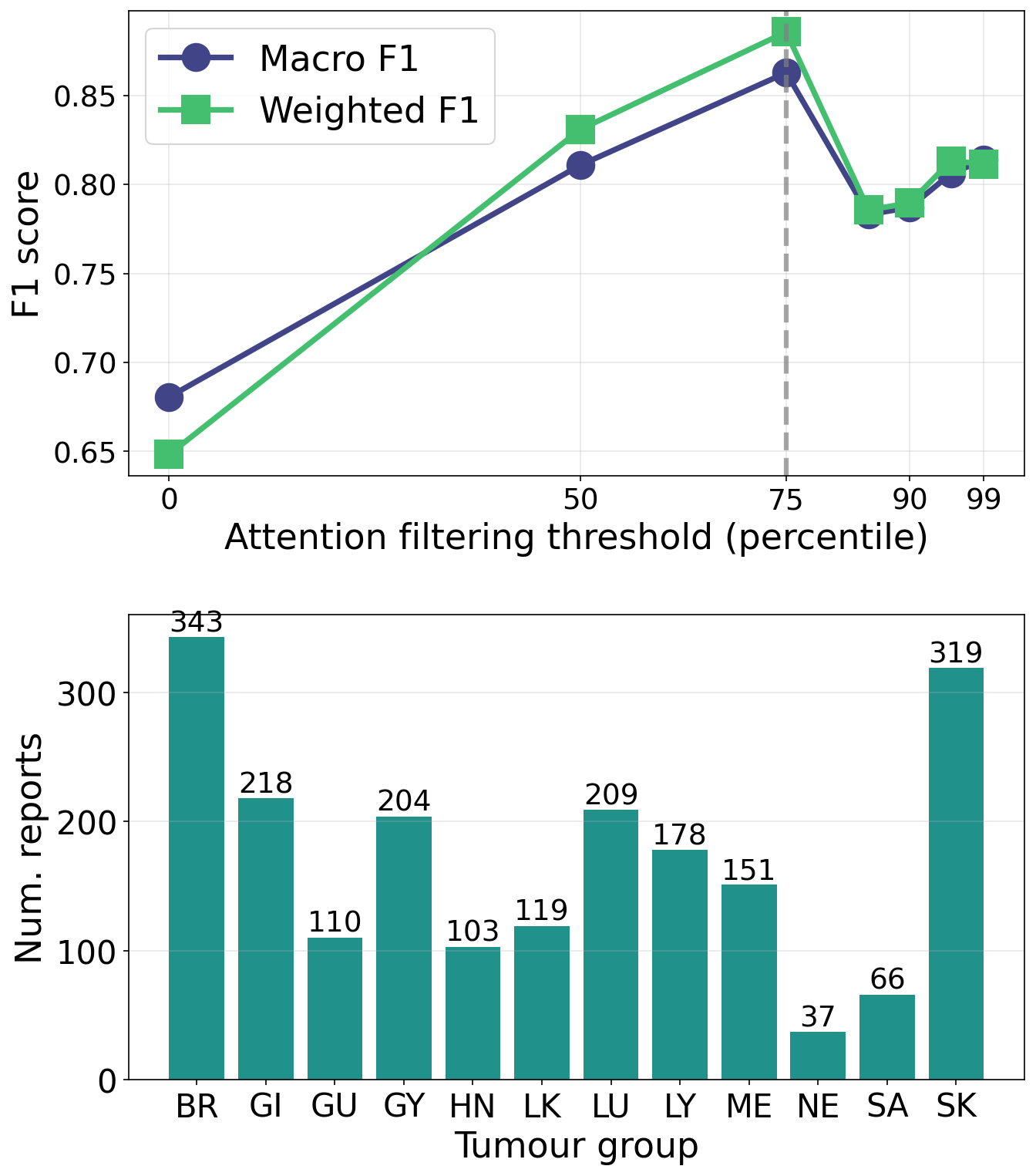}
\caption{ABMIL-DISTIL classifier performance as a function of attention filtering threshold on half of the ODS-annotated dataset (dev). (Top) Macro and weighted F1 across thresholds (0th–99th percentile). The dashed line marks the threshold selected via two-fold cross-validation on the audit set. (Bottom) Per-class support (number of reports per class). Rarer tumour groups (notably NE, n=37, and SA, n=66) performance is estimated on substantially fewer examples.}
		\label{f1_support_plots}
\end{figure}

Next the ABMIL-DISTIL stage is run on a set of filtered datasets. Figure \ref{radar_threshold_comparison} shows macro precision, recall, and accuracy as a function of attention score percentile threshold across tumour groups, with corresponding aggregate F1 scores in Figure \ref{f1_support_plots}. At the 0\% threshold — equivalent to the label-propagation baseline, where every report is retained and assigned the propagated PHN-level label — performance was worst. Macro F1 improved substantially as the threshold increased to 75\%, gains attributable to improved training signal quality rather than any architectural change, since the only difference is the removal of low-attention reports. Beyond 75\%, macro F1 declined, consistent with the trade-off between label quality and training set size: at high thresholds the reduction in training data outweighs the gain in label cleanliness, particularly for minority tumor groups with limited representation in the filtered corpus. The 75\% threshold was therefore selected as optimal and used for the end-to-end SFT run described below.

\begin{figure}[t]
\centering
\includegraphics[width=0.7\textwidth]{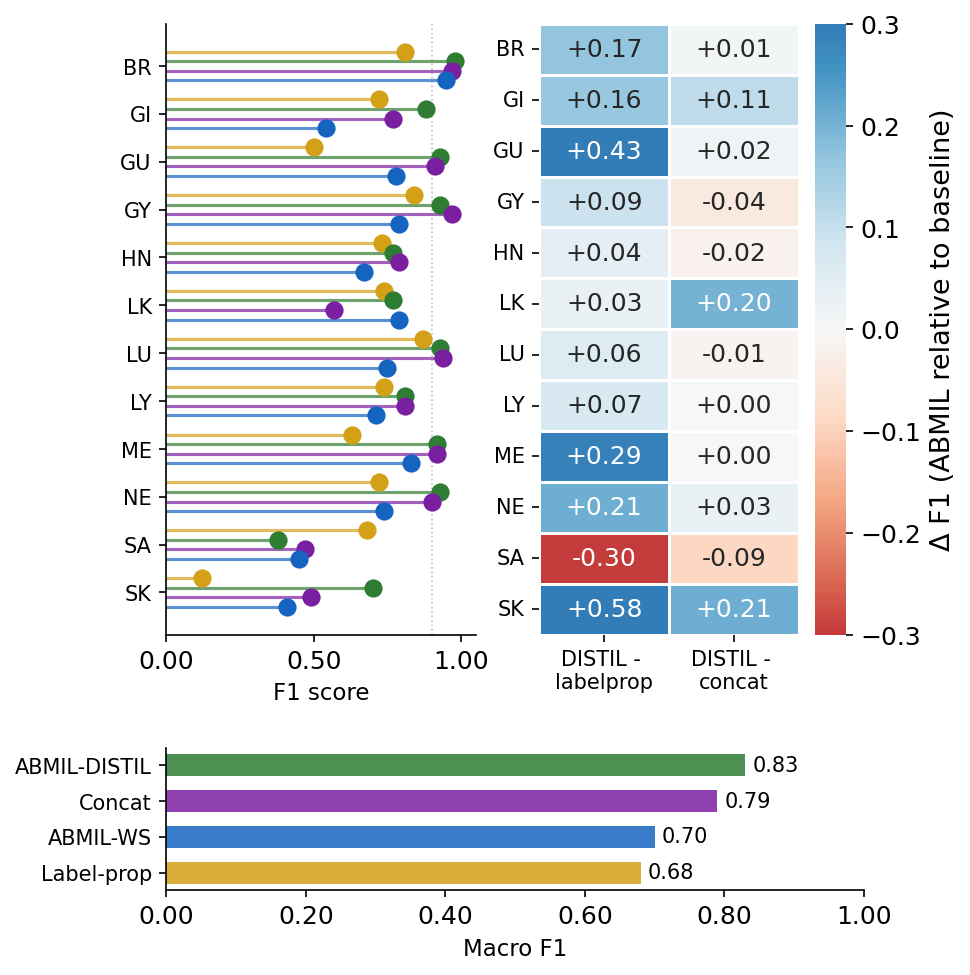}
\caption{Per-class F1 scores on the test half of the ODS-annotated dataset for four methods: label propagation (yellow), ABMIL-DISTIL at 75\% threshold (green), concatenation baseline (purple), and ABMIL-WS (blue). The dot plot (left) shows absolute F1 per tumour group. The heatmap (right) shows the per-class change in F1 between the ABMIL-DISTIL model relative to label propagation (left column) and the concatenation baseline (right column). The bottom plot reports Macro-F1 per model.}
		\label{model_comparison_results}
\end{figure}

Having demonstrated performance change as a function of threshold using the ABMIL-DISTIL pipeline and selected an optimum threshold (75\%), we now contextualize performance relative to the \textit{label-propagation} and \textit{concatenation-baselines}. In Figure \ref{model_comparison_results}, the ABMIL-DISTIL pipeline is shown to achieve the strongest absolute performance with a macro-F1 of 0.83, outperforming the \textit{label propagation baseline} (macro F1 = 0.68) and the \textit{concatenation baseline} (macro F1 = 0.79) across the majority of tumor groups. Per-class F1 gains over label propagation are consistent and broad, with the largest improvements for Skin, Genitourinary and Melanoma. The margin over the concatenation baseline is narrower, reflecting the competitive performance of full-sequence attention over a large context window. 

The principal exception is the Sarcoma group, for which ABMIL-DISTIL (0.38) significantly underperforms label propagation (0.68). The confusion matrices (Supplementary Material) show this is not a failure of sarcoma recall — which is high (0.89) — but of precision (0.25): sarcoma functions as a partial catch-all class, absorbing misclassified reports from other groups, most prominently skin. Sarcomas comprise over 50 subtypes with overlapping, hard-to-distinguish features \cite{vandijke2009soft}; combined with limited training data, this may produce a diffuse embedding region that absorbs ambiguous reports — consistent with the catch-all behavior observed here.

Finally, the ABMIL-WS stage tested as a standalone classifier achieved a macro F1 of 0.72 on the ODS-annotated dataset, slightly outperforming the label-propagation baseline. As the ODS dataset is labeled at the report level, each report is treated as a single-instance bag at inference.

\section{Discussion}
This study presents the first application of ABMIL to tumor group classification from pathology reports using operationally-generated PHN-level supervision. The approach uses attention weights as a data curation mechanism: applying a percentile-based filtering threshold to the corpus-wide attention score distribution yielded progressively cleaner training datasets, with classification performance peaking at the 75th percentile threshold. These results establish ABMIL as a principled, computationally efficient, and practically deployable framework for leveraging operational registry data for cancer NLP, one that requires no additional annotation effort beyond what registries already produce in the course of their routine operations.

The ABMIL-DISTIL model trained on an attention-filtered dataset achieved the strongest performance of the methods evaluated. Given the modest label noise introduced by the mapping schema discrepancy described above, this result should be interpreted as a conservative estimate of true model performance. Performance is likely further constrained by the partial unfreezing strategy adopted for computational tractability, suggesting that the reported figures represent a floor rather than a ceiling on achievable accuracy. We caution that the 75th percentile threshold is not a universal optimum but an artifact of this corpus and its noise characteristics. Datasets constructed under different filtering or inclusion criteria, leading to different label noise profiles, would likely require their own bespoke threshold tuning, and the threshold should be treated as a dataset-specific hyperparameter rather than a transferable constant.


Additionally, the ABMIL-WS classifier was found to marginally exceed the performance of the label propagation baseline despite occupying a fundamentally different computational tier. It requires no encoder fine-tuning and no GPU for training the attention head. Once report embeddings have been extracted, the full pipeline can in principle run on a CPU alone. That a method requiring only frozen embeddings and a lightweight attention head matches or exceeds a baseline requiring full per-report fine-tuning is a meaningful result for resource-constrained registries: for institutions without GPU infrastructure, this represents a viable entry point into automated tumor group classification that no other method evaluated here can match. 

Several limitations of this study should be acknowledged. First, all experiments were conducted on data from a single registry and the generalizability of the findings to other PBCRs with different coding practices, pathology report formats, and tumor group taxonomies remains to be established. Secondly, the ABMIL framework implemented here assumes a single tumor group label per PHN, enforced by the single-diagnosis filtering step described in the methods. Patients with multiple primary tumors, a non-trivial proportion of cancer registry populations, are excluded from training and evaluation. Finally, the attention mechanism assumes that at least one report within each bag is relevant to the assigned tumor group. In practice this may not hold: ODS decisions can draw on imaging reports, surgical notes, and other clinical documentation beyond the pathology record, so the evidence supporting a given label may be absent from the bag entirely. We adopted the standard ABMIL formulation in this work, and leave relaxation of this assumption as a direction for future work.

A broader insight emerging from these results is that \textit{attention} is the operative mechanism in both of the most successful approaches in this study. Whether applied across a dissolved sequence of concatenated reports or at the report level via ABMIL pooling, the model is in both cases learning to parse diagnostically relevant signal from a noisy collection of documents and solve the lost provenance problem. The two baselines used in this study are each computationally demanding in a different way — label propagation scales with the full corpus size, while concatenation scales with sequence length — whereas the ABMIL filtering pathway sidesteps both by reducing the corpus to a small set of high-attention reports before fine-tuning. The ABMIL framework preserves the report as the fundamental unit, enabling an intermediate model to be trained efficiently on frozen embeddings to identify the most relevant reports, after which the filtered dataset is small enough that end-to-end fine-tuning becomes computationally tractable on commercial-grade GPUs. The fact that both methods achieve comparable performance on the ODS-annotated dataset, with the ABMIL filtering pipeline even outperforming the concatenation baseline, suggests that the dataset distillation pathway is a practically superior solution to the granularity gap problem that all PBCRs face.

While this study focuses on tumor group classification as the target endpoint, the ABMIL framework is not inherently task-specific and the dataset distillation pathway generalizes in principle to any registry-derived patient-level label for which a linked pathology report corpus exists. PBCRs routinely code a broad range of tumor attributes beyond diagnostic group, including histology, primary site, laterality, behavior, and pathological staging, all of which are recorded at the patient or tumor level in cancer registries. Each of these constitutes a potential bag-level supervision signal that could be exploited within the same framework: a separate ABMIL model trained on each label type would learn to concentrate attention on the reports most relevant to that specific coding decision, generating a filtered dataset suitable for downstream SFT.

\bibliographystyle{vancouver}
\bibliography{references}
\end{document}